\pdfoutput=1

\documentclass[11pt]{article}

\usepackage[preprint]{acl}

\usepackage{times}
\usepackage{latexsym}

\usepackage[T1]{fontenc}

\usepackage[utf8]{inputenc}

\usepackage{microtype}

\usepackage{inconsolata}

\usepackage{graphicx}

\usepackage{amsmath}
\usepackage{enumitem}
\setlist[itemize]{itemsep=2pt, parsep=2pt, topsep=3pt, partopsep=0pt}
\usepackage{graphicx}
\usepackage{booktabs}
\usepackage{array}
\usepackage{graphicx}
\usepackage{listings}
\usepackage{tabularx}  
\usepackage{tablefootnote}  
\usepackage{caption}  
\usepackage{subcaption}
\usepackage{xcolor}

\definecolor{codegreen}{rgb}{0,0.6,0}
\definecolor{codegray}{rgb}{0.5,0.5,0.5}
\definecolor{codepurple}{rgb}{0.58,0,0.82}
\definecolor{backcolour}{rgb}{0.95,0.95,0.92}
\lstdefinestyle{mystyle}{
    backgroundcolor=\color{backcolour},   
    commentstyle=\color{codegreen},
    keywordstyle=\color{magenta},
    numberstyle=\tiny\color{codegray},
    stringstyle=\color{codepurple},
    basicstyle=\ttfamily\footnotesize,
    breakatwhitespace=false,         
    breaklines=true,                 
    captionpos=b,                    
    keepspaces=true,                 
    numbers=left,                    
    numbersep=5pt,                  
    showspaces=false,                
    showstringspaces=false,
    showtabs=false,                  
    tabsize=2,
    inputencoding=utf8,
    escapeinside={(*@}{@*)}, 
}
\usepackage{multirow}  
\usepackage{tikz,tcolorbox}
\usepackage{algorithm}
\usepackage{algpseudocode}
\usepackage{amsthm}
\usepackage{makecell}
\usepackage{mathtools}
\usepackage[algo2e]{algorithm2e}

\usepackage{amsmath,amsfonts,bm}

\def\eqref#1{equation~\ref{#1}}

\def\1{\bm{1}}

\def\rvx{{\mathbf{x}}}
\def\rvy{{\mathbf{y}}}

\DeclareMathAlphabet{\mathsfit}{\encodingdefault}{\sfdefault}{m}{sl}
\SetMathAlphabet{\mathsfit}{bold}{\encodingdefault}{\sfdefault}{bx}{n}

\newcommand{\KL}{D_{\mathrm{KL}}}

\title{$\mathcal{B}^4$: A \underline{B}lack-\underline{B}ox Scru\underline{B}\underline{B}ing Attack on LLM Watermarks}

\newcommand{\affilPKU}{\ensuremath{^\clubsuit}}
\newcommand{\affilUCSB}{\ensuremath{^\diamondsuit}}

\author{
Baizhou Huang \affilPKU\thanks{Equal contribution.} \quad
Xiao Pu \affilUCSB$^*$ \quad
Xiaojun Wan \affilPKU \\
\affilPKU Peking Univerisity \quad
\affilUCSB University of California, Santa Barbara \\
{\tt \{hbz19,wanxiaojun\}@pku.edu.cn  \quad xiao\_pu@ucsb.edu}
}

\begin{document}
\maketitle
\begin{abstract}
Watermarking has emerged as a prominent technique for LLM-generated content detection by embedding imperceptible patterns. 
Despite supreme performance, its robustness against adversarial attacks remains underexplored.
Previous work typically considers a grey-box attack setting, where the specific type of watermark is already known. Some methods even necessitates knowledge about details of hyperparameters. Such prerequisites are unattainable in real-world scenarios.
Targeting at a more realistic black-box threat model with fewer assumptions, we here propose $\mathcal{B}^4$, a \textbf{B}lack-\textbf{B}ox scru\textbf{BB}ing attack on LLM watermarks. Specifically, we formulate the watermark scrubbing attack as a constrained optimization problem by capturing its objectives with two distributions: a \textit{Watermark Distribution} and a \textit{Fidelity Distribution}. The optimization problem can be approximately solved using two proxy distributions.
Experimental results across 12 different settings demonstrate the superior performance of $\mathcal{B}^4$ compared with other baselines. 
\footnote{The code will be released after anonymous reviews.}
\end{abstract}

\newcommand{\secvs}{\vspace{-2.5mm}}
\newcommand{\subsecvs}{\vspace{-2mm}}
\newcommand{\figvsbottom}{\vspace{-3mm}}
\newtheorem{corollary}{Corollary}
\newtheorem{proposition}{Proposition}[section]
\newtheorem{lemma}{Lemma}[section]
\newcommand{\Wdist}{\ensuremath{{P_w}}}
\newcommand{\hWdist}{\ensuremath{{\hat{P_w}}}}
\newcommand{\Fdist}{\ensuremath{{P_f}}}
\newcommand{\hFdist}{\ensuremath{{\hat{P_f}}}}

\section{Introduction}
\subsecvs


\begin{figure}[t]
\centering
\includegraphics[width=0.8\linewidth]{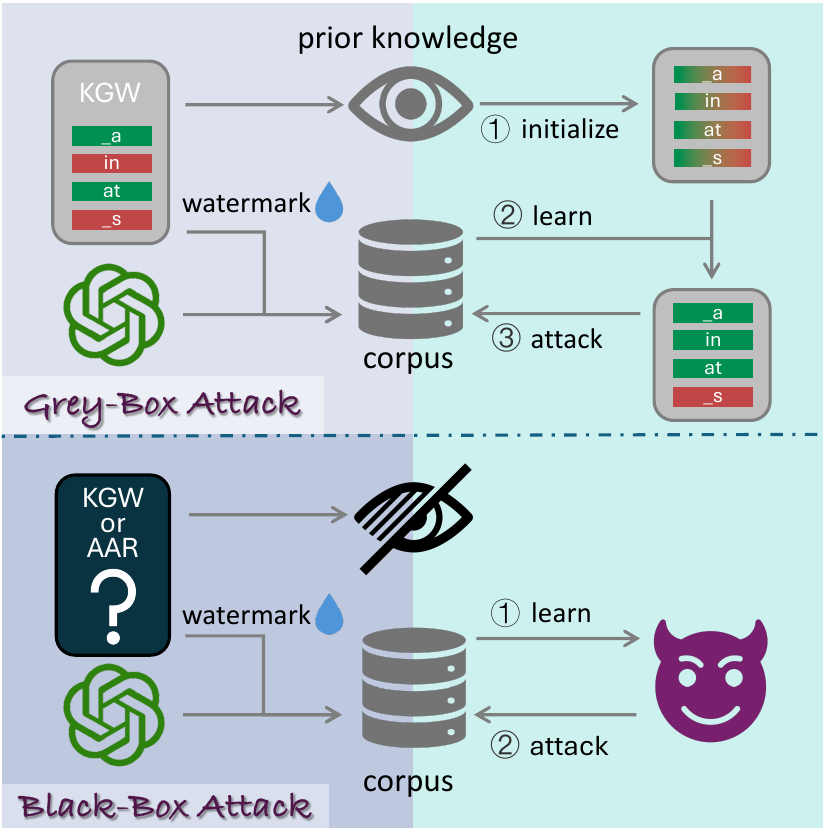}
\vspace{-1mm}
\caption{Difference between grey-box and black-box threat models. The left part in purple represents the victim and the right part in green represents the attacker. (Top) Prior work used prior knowledge for parametrization, e.g. the green-list partition of vocabulary in KGW, which makes watermark stealing easier.
(Bottom) Under a more realistic black-box setting, the problem becomes much more challenging.}
\figvsbottom
\end{figure}

The rapid advancement of large language models (LLMs) has demonstrated their unimaginable potentials across various applications. Systems such as ChatGPT \citep{openai2023gpt4} are now seamlessly integrated into many aspects of daily life. Despite benefits, the extensive deployment of LLMs has sparked serious concerns regarding potential misuse, such as large-scale disinformation, automated spamming, and social media manipulation, thereby threatening academic integrity and intellectual property rights \cite{10.1145/3442188.3445922,liuSurveyTextWatermarking2023}.
Consequently, detecting LLM-generated content has emerged as a crucial focus in the discourse on LLM safety and responsible deployment \cite{mitchell2023detectgptzeroshotmachinegeneratedtext,pu-etal-2023-zero,yangSurveyDetectionLLMsGenerated2023}. 

Watermarking stands out as a prominent technique for detecting LLM-generated text. It injects a hidden pattern invisible to human into generated contents of a specific LLM \cite{kirchenbauerWatermarkLargeLanguage2023}. 
By altering the original distribution of LLMs to a specific watermark distribution during each decoding step \cite{kuditipudiRobustDistortionfreeWatermarks2023,zhaoProvableRobustWatermarking2023,huUnbiasedWatermarkLarge2023}
, all model generated contents can be statistically distinguished through hypothesis testing between the watermarked and the original distributions.
This approach achieves a high detection rate and can be easily deployed, with only a negligible cost in the quality of generated content.

Despite its supreme performance, the robustness of watermarking methods against adversarial attacks remains underexplored. Among which, the \textit{scrubbing attack} \cite{jovanovicWatermarkStealingLarge2024} presents a notable challenge in practical settings: if an adversary can successfully paraphrase LLM-generated contents into another semantically similar but watermark-free form, the effectiveness of the watermark will be heavily compromised.

\citet{wuBypassingLLMWatermarks2024} and \citet{jovanovicWatermarkStealingLarge2024} explored this question by proposing different watermark scrubbing methods. However, these approaches typically require strong prior knowledge about the specific type of watermark algorithm used in the victim LLM, even the specific hyperparameter of the algorithm (e.g. the context window size in KGW). 
For example, \citet{wuBypassingLLMWatermarks2024} proposed a method based on the assumption that the victim LLM is protected with KGW watermarking algorithm \citep{kirchenbauerWatermarkLargeLanguage2023}, so that they can parameterize their attack model based on the green-red list partition. 
Such prerequisites are virtually unattainable in real-world scenarios, especially when we can only access the victim LLM via an API interface. The stringent constraint on prior information makes these attacking methods impractical for real-world applications.

In this work, we consider a more realistic black-box threat model with the sole prior knowledge of watermark existence.
Under this attack setting, we further propose $\mathcal{B}^4$ (\textbf{B}lack-\textbf{B}ox scru\textbf{BB}ing attack), a universal watermark attack method, specially designed for the black-box threat model. With less assumption about the victim LLM, this method is more practical for real-life usage, allowing us to accurately explore the robustness of current watermark techniques. 

The intuition of $\mathcal{B}^4$ is rooted in two fundamental objectives of an ideal watermark scrubbing attack: the adversarial texts should both exhibit minimal watermark patterns to evade from detection (\textit{Efficacy}) and preserve the semantic information of the original texts (\textit{Fidelity}). Drawing on this insight, we propose to capture these dual properties with distance to two distributions respectively: a \textit{Watermark Distribution} and a \textit{Fidelity Distribution}. Consequently, we formulate the task of watermark scrubbing attack into a constrained optimization problem. The local optimal of this problem can be solved by approximating these distributions with two proxies.
Specifically, we steer a much smaller proxy model to approximate the \textit{Watermark Distribution} of the victim LLM through distillation, and apply a general paraphrase model to obtain a distribution similar to the original texts for the \textit{Fidelity Distribution}. 

Compared to baseline watermark srubbing methods, our proposed $\mathcal{B}^4$ improves the attack success rate by up to 68.13\% across different victim LLMs and watermarking method settings. 
Since there is an inherent trade-off between attack efficacy and semantic fidelity, we further demonstrate the superior performance of $\mathcal{B}^4$ over baselines using Pareto fronts: $\mathcal{B}^4$ removes text watermarks more efficiently under the same fidelity constraints, while also distorting less semantic information to achieve the same level of attack.

In summary, our contributions include: (1) formalizing the watermark scrubbing attack as a constrained optimization problem for the first time, and (2) developing a black-box watermark scrubbing method based on this formulation, which demonstrates superior performance compared to previous approaches.

\secvs
\section{Preliminary}
\subsection{LLM Watermarking}
\label{sec:prelim-watermark}
We begin by introducing LLM watermarking. Given any prompt $\rvx$, an LLM generates a sequence of output tokens $\rvy_i\sim P(\cdot|\rvx,\rvy_{<i})$ in an auto-regressive manner.
Watermarking modifies this distribution into a distorted distribution, $\Wdist(\cdot |\rvx,\rvy_{<i})$, embedding hidden patterns associated with a secret key.
Detecting watermarked text can thus be framed as a hypothesis testing problem, with the alternative hypothesis positing that the candidate text is sampled from the altered distribution. Typically, this detection process involves accumulating per-token statistics $s_i$ to conduct a one-tailed significance test.

Initial research on LLM watermarking \citep{aaronson-watermark,kirchenbauerWatermarkLargeLanguage2023} demonstrated promising results, inspiring numerous follow-up works that introduced diverse watermarking algorithms \cite{kirchenbauerReliabilityWatermarksLarge2023,kuditipudiRobustDistortionfreeWatermarks2023,zhaoProvableRobustWatermarking2023,huUnbiasedWatermarkLarge2023,liuSemanticInvariantRobust2023,houSemStampSemanticWatermark2023}. In this paper, we focus on four prominent watermarking techniques: KGW\cite{kirchenbauerWatermarkLargeLanguage2023}, AAR \cite{aaronson-watermark}, Unigram \cite{zhaoProvableRobustWatermarking2023} and EXP \cite{kuditipudiRobustDistortionfreeWatermarks2023}. We provide a detailed description of these watermarking methods in Appendix \ref{app:watermarks}.

\subsection{Watermark Attack Threat Model}
\label{sec:prelim-task}
An ideal watermarking algorithm should not only be able to precisely distinguish watermarked texts, but also be resilient to adversarial attacks. The watermark scrubbing attack can be viewed as a specialized form of post-hoc paraphrasing, where the original \textit{Watermarked Sample} $\rvy^w$ is transformed into  \textit{Adversarial Sample} $\rvy$. It should retain similar semantics (\textit{Fidelity}) while evading detection by removing watermark patterns (\textit{Efficacy}). 

In this work, our threat model consists of two main actors: the victim and the attacker. The \textit{victim} is a proprietary large language model, protected by a watermarking algorithm and accessible only through an API service interface. The \textit{attacker} is able to query the victim model with different prompts through the interface to obtain corresponding responses. We only make the minimal assumption that the attacker is informed with the existence of watermarks but possesses no prior knowledge of the specific type of the watermarking algorithm used. This stands in contrast to previous work that commonly relies on strong assumptions, such as the details of hyperparameters \cite{wuBypassingLLMWatermarks2024,jovanovicWatermarkStealingLarge2024} or an oracle detector \cite{pangAttackingLLMWatermarks2024}, which severely restrict the range of attackable watermarking algorithms.

Our threat model is more closely aligned with real-world scenarios -- Most leading companies such as OpenAI and Anthropic provide online API-based services under similar settings. By adhering to this realistic threat model, our investigation will provide more meaningful insights into the robustness of current watermarking techniques, fostering a deeper understanding of their vulnerabilities.

\secvs
\section{Method}
In this section, we introduce $\mathcal{B}^4$, a universal watermark attacking method under the black-box threat model settings. We begin by explaining the intuition behind the method, followed by formalizing watermark scrubbing attack as an optimization problem. Next, we detail the numerical solving process of the formulated problem via two proxy distributions. Finally, we conclude with a discussion of potential approximation errors, along with a proposed adjustment for alleviation. We present the pseudocodes of our method in Appendix \ref{app-pcodes}.
\subsecvs
\subsection{Scrubbing as an Optimization Problem}
\label{sec:method-problem}
As mentioned in Section \ref{sec:prelim-watermark}, various watermarking algorithms differ in their watermark injection and detection mechanisms, which often causes a scrubbing attack effective against one algorithm but failing against others. Nonetheless, all watermarking techniques fundamentally involve altering the token distribution, regardless of the specific parametrization or detection statistics used. Thus, we can generalize the watermarking process as a transformation to a watermark distribution, $\Wdist(\rvy)$

Suppose an ideal adversarial sample $\rvy$ of watermark attack is drawn from a probability distribution $Q(\cdot|\rvy^w)$, conditioned on a watermarked sample $\rvy^w$.
We can then characterize the objectives of watermark scrubbing attack using mathematical expressions. Fidelity reflects the similarity between $\rvy$ and $\rvy^w$, which can be expressed by the divergence from an implicit \textit{Fidelity Distribution} $\Fdist(\rvy|\rvy^w)\propto \text{SIM}(\rvy,\rvy^w)$, where $\text{SIM}(\cdot,\cdot)$ is a true similarity measure between two texts. Efficacy, on the other hand, requires to minimize the likelihood that a watermark detector successfully identifies the adversarial sample as being watermarked. One sufficient condition to achieve this is to maximize the distance from the \textit{Watermark Distribution} $\Wdist$, since the watermark detection is a hypothesis testing with $\Wdist$ as the distribution under the alternative hypothesis. Therefore, to find the optimal solution $Q$ can be formalized into the following constrained optimization problem:
\vspace{-2mm}
{\thinmuskip=0.5mu
\medmuskip=2mu
\thickmuskip=2mu
\begin{equation}
    \label{eq:problem}
    \min_Q -\KL(Q||\Wdist)\quad s.t.\,\KL(Q||\Fdist)\leq\epsilon
\end{equation}}
where $\epsilon$ is a hyperparameter controlling the degree of allowed semantic deviation from the original watermarked sample. Since the problem satisfies the Slater Constraint Qualification \cite[Proposition 3.3.9]{bertsekas1999nonlinear}, the local minimal are guaranteed to satisfy the Karush-Kuhn-Tucker (KKT) conditions. Solving the KKT conditions for the problem above immediately yields the following corollary:
\vspace{-1mm}
\begin{corollary}
\label{collary}
    The local minimum point $Q^*$ has the form of 
    \vspace{-2mm}
    \begin{equation}
    \label{eq:optimal-expresion}
        Q^*(\rvy|\rvy^w)=\frac1Z\Fdist^{\frac{1}{1-\lambda^*}}(\rvy|\rvy^w)\Wdist^{-\frac{\lambda^*}{1-\lambda^*}}(\rvy)
    \end{equation}, where $\lambda^*\in (0,1)$ is the corresponding Lagrange multiplier satisfying $\KL(Q^*||\Fdist)=\epsilon$, and $Z$ is the normalizing constant. 
\end{corollary}
\vspace{-1mm}
The corresponding Lagrange multiplier $\lambda^*$ can be solved using the Newton-Raphson Method. 

\subsecvs
\subsection{Approximated Solution with Proxy Distributions}
\subsecvs
To solve the optimization problem above, we also need to parameterize both $\Wdist$ and $\Fdist$. However, both of them are inherently intractable. Given the lack of prior knowledge regarding the watermarking algorithm behind the API, obtaining the accurate form of $\Wdist$ is impossible. Therefore, we leverage model distillation \cite{hinton2015distillingknowledgeneuralnetwork} to learn a proxy, denoted as $\hWdist$ \footnote{Throughout this work, proxy distributions are denoted using the hat notation ($\hat~$).}. In specific, we sample from the golden watermark distribution by querying the victim LLM for multiple times. Those responses form a training corpus $\mathcal D$ to train a local language model $p_\theta$ by minimizing the Negative Log-Likelihood (NLL) loss,
\vspace{-2mm}
{\thinmuskip=1mu
\medmuskip=1.5mu
\thickmuskip=1.5mu
\begin{equation*}
    \mathcal{L}(\theta) = -\sum_{\rvy\in \mathcal D}\log \hWdist(\rvy;\theta)=-\mspace{-10mu}\sum_{\rvy\in \mathcal D;i}\log p_\theta(\rvy_i|\rvy_{<i})
\end{equation*}}
\vspace{-1mm}
The choice of the local model can vary among numerous open-source LLMs available. 
Empirically, we find that a proxy watermark distribution can be learned surprisingly well from a moderately sized corpus.

As for the fidelity distribution $\Fdist$, we simply apply a paraphrase model $p_\phi$ as the proxy fidelity distribution, with the watermarked sample $\rvy^w$ serving as context, i.e. $\hFdist(\rvy|\rvy^w;\phi)=\prod_i p_\phi(\rvy_i|\rvy_{<i}, \rvy^w)$.
Now that we are able to parameterize the local optimal $Q^*$, transforming Equation \ref{eq:optimal-expresion} into:
\vspace{-1mm}
\begin{equation}
\label{eq:optimal-expression-AR}
    Q^*(\rvy_i|\rvy_{<i},\rvy^w)=\frac{\hFdist^{\frac{1}{1-\lambda^*}}(\rvy_i|\rvy_{<i},\rvy^w;\phi)}{\hWdist^{\frac{\lambda^*}{1-\lambda^*}}(\rvy_i|\rvy_{<i};\theta)}
\end{equation}
This allows us to sample adversarial texts in an auto-regressive manner\footnote{We ignore the normalizing constant for simplicity.}.





\subsecvs
\subsection{Approximation Error Adjustment (AEA)}
\label{sec:aea}
\subsecvs
While proxy distributions provide a practical approximation for solving the optimization problem, they may diverge from the golden distributions in certain regions of sample space. 
This issue is particularly severe for the proxy watermark distribution $\hWdist$, due to the inherent limitations of sampling-based model distillation. 

Unlike logit-based distillation \cite{hinton2015distillingknowledgeneuralnetwork}, which offers a holistic guidance over the entire space, sampling-based distillation \citep{kim-rush-2016-sequence} applies a one-peak correction on certain regions around each observed sample by the NLL loss. 
When the training data is sparse, the student model may struggle to generalize to unobserved regions of the teacher watermark distribution $\Wdist$. For instance, in the Unigram algorithm, green list tokens generally have higher output probabilities. However, due to the randomness of sampling, some of them may remain unobserved, leading to significant discrepancies between their probabilities in the proxy distribution $\hWdist$ and the golden distribution $\Wdist$. These approximation errors may critically affect our method, given that the optimization objective involves minimizing KL divergence, a global measure over the entire sample space.

To address this issue, we introduce an adjustment mechanism to refine the approximation errors by simply excluding those under-fitting regions from the calculation of the KL divergence objective. 
We identify these regions, $\Sigma_u^i$, which is a subset of vocabulary $\Sigma$, by comparing the proxy distributions before and after distillation at each decoding step, given the context $\rvy_{<i}$:
\begin{equation*}
    \Sigma_u^i= \left\{v\in \Sigma: \left|p_\theta(v|\rvy_{<i})-p_{\theta_{ini}}(v|\rvy_{<i})\right|<\mu\right\}
\end{equation*}
where $\theta_{ini}$ is the initialized weights before distillation, and $\mu$ is a pre-defined threshold\footnote{We apply the uniform probability $1/|\Sigma|$ as threshold.}. 
Eventually, we can incorporate the approximation error adjustment into Equation \ref{eq:optimal-expression-AR} to obtain the adjusted attack distribution:
\begin{equation}
\label{eq:optimal-expression-final}
\small
\thinmuskip=1mu
\medmuskip=1.5mu
\thickmuskip=1.5mu
    Q^*(\rvy_i|\rvy_{<i},\rvy^w)=
    \begin{cases}
        \hFdist(\rvy_i|\rvy_{<i},\rvy^w;\phi) & \text{if } \rvy_i\in\Sigma_u^i,\\
        \frac{\hFdist^{\frac{1}{1-\lambda^*}}(\rvy_i|\rvy_{<i},\rvy^w;\phi)}{\hWdist^{\frac{\lambda^*}{1-\lambda^*}}(\rvy_i|\rvy_{<i};\theta)} & \text{otherwise}.
    \end{cases}
\end{equation}

Besides the above adjustment, the proposed $\mathcal{B}^4$ is also compatible with all existing decoding strategies, such as nucleus sampling \cite{holtzman2020curiouscaseneuraltext}. 
In this paper, we apply both top-$50$ and $10$-beam search to ensure high-quality outputs.



\secvs
\section{Experiments}
\subsecvs
\subsection{Experimental Settings}
\subsubsection{Victim Models and Target Datasets} 
In order to simulate different scenarios in the real world, we consider three different sizes of victim LLMs from two well-known model families, including Llama-2-7B \cite{llama1}\footnote{\url{https://huggingface.co/meta-llama/Llama-2-7b}}, Llama-3-70B \cite{llama2}\footnote{\url{https://huggingface.co/meta-llama/Meta-Llama-3-70B}} and Qwen2-72B \cite{qwen2}\footnote{\url{https://huggingface.co/Qwen/Qwen2-72B}}.
We implement four watermarking algorithms mentioned in Section \ref{sec:prelim-watermark}, i.e. KGW, Unigram, EXP and AAR.
We follow the experimental setting in previous work \cite{kirchenbauerWatermarkLargeLanguage2023,kirchenbauerReliabilityWatermarksLarge2023} to utilize a random subset of C4-Realnewslike dataset \citep{2020t5} as prompts to query these watermarked models for 100 responses in the length of 200 tokens. These responses construct the watermarked sample dataset to evaluate different attack methods.
\subsecvs
\subsubsection{Baselines}
Given the black-box threat model setting, we consider three watermark attack baselines in our experiments.
\textbf{Base} directly utilizes an LLM to paraphrase the watermarked sample via prompting. We apply the identical paraphraser $\phi$ used by $\mathcal{B}^4$ for fair comparison. Recursive paraphrasing (\textbf{RP}) applies a paraphraser to paraphrase the watermarked sample in chunk-level for multiple times \cite{sadasivanCanAIGeneratedText2023} . We also use the identical paraphraser $\phi$ for fair comparison. 
\textbf{DIPPER} \cite{krishnaParaphrasingEvadesDetectors2023} is a T5-XXL-based paraphraser specifically tuned for evading AI-generation detection. 
\vspace{-1mm}
\subsubsection{Metrics}
\label{sec:exper-metric}
\vspace{-1mm}
As stated in Section \ref{sec:prelim-task}, \textit{Fidelity} and \textit{Efficacy} serve as two goals of our watermark scrubbing method.
For fidelity, we follow \citet{krishnaParaphrasingEvadesDetectors2023,jovanovicWatermarkStealingLarge2024} to use P-SP \cite{wieting-etal-2022-paraphrastic} to reflect the semantic similarity between the texts before and after attacks. 
For efficacy, we use 
ROC-AUC \citep{hanley1982meaning} to evaluate the watermark strength after attacks. 
One should be noted that there is an inherent trade-off between these two metrics: under smaller fidelity constraint, the valid semantics space is more limited for modification to remove the watermark patterns within the original text, thereby inevitably leading to lower efficacy.
Due to this, instead of comparing different methods within a single dimension, we aim to identify an attacking method that is optimal in terms of both fidelity and efficacy. We borrow the concept of Pareto front from economics as an evaluation method, which consists of a group of optimal states in multi-objective scenarios. In practice, we apply each method to generate diverse adversarial samples under different fidelity constraints by grid searching its hyperparameter space. 
Take $\mathcal{B}^4$ as an example, we vary $\epsilon$, the fidelity constraint threshold defined in Problem \ref{eq:problem}, to obtain diverse data points with different fidelity and efficacy, thereby forming a Pareto front curve as shown in Figure \ref{fig:main}. The hyperparameter space of other baselines are presented in Appendix \ref{app-setting}.
For a more intuitive comparison, we also include the numerical AUC@$f$ metrics, which indicates the best ROC-AUC score that an attacking method can achieve under a specific fidelity constraint $f$.

\begin{figure*}[!htb]
    \centering
    \includegraphics[width=\linewidth]{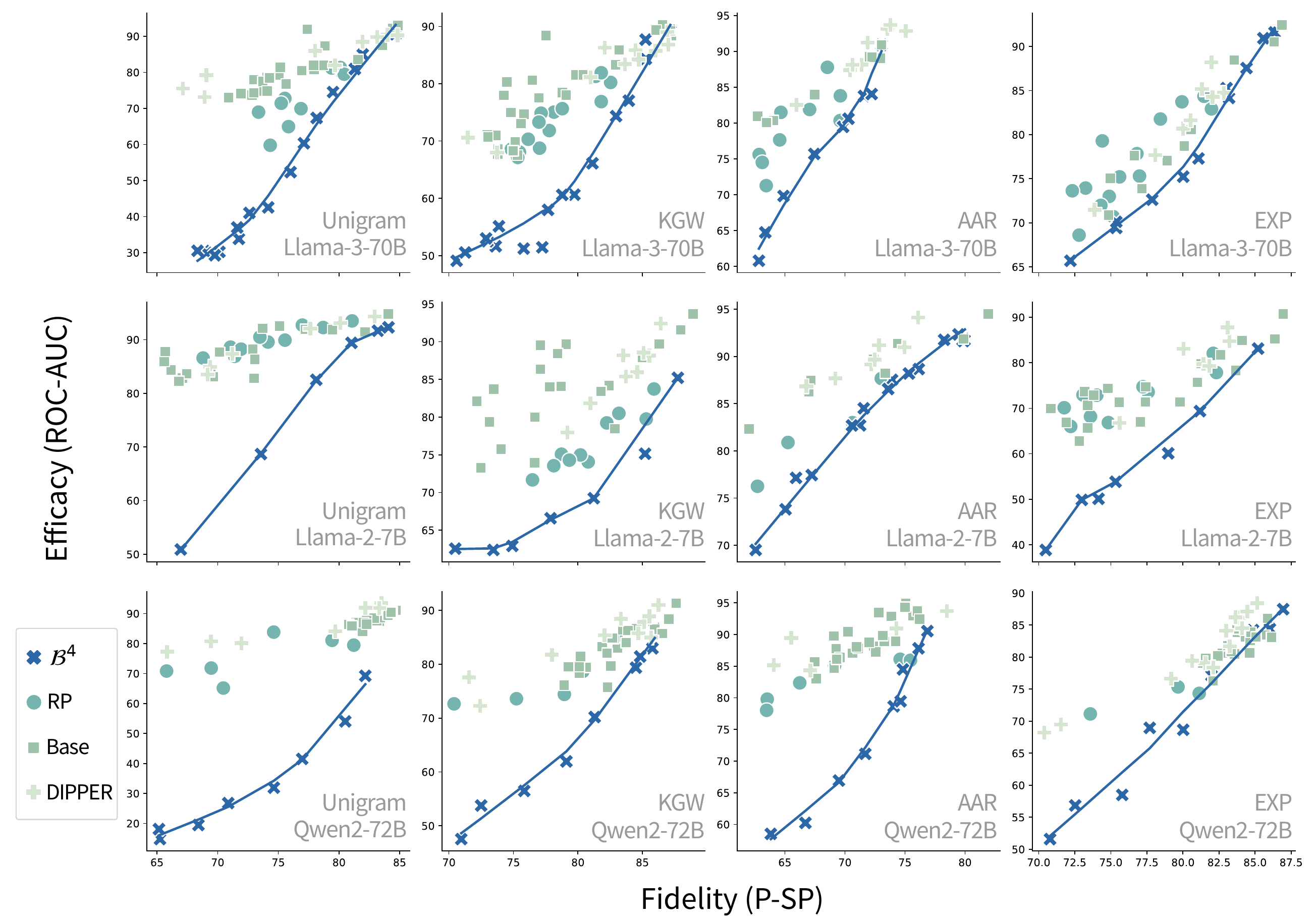}
    \figvsbottom
     \caption{Performance of watermark scrubbing attack methods against different victim LLMs protected by different watermark algorithms. The $y$-axis indicates Efficacy, measured by ROC-AUC ($\downarrow$), while the $x$-axis indicates Fidelity, measured by P-SP ($\uparrow$). Each data point represents one watermarking algorithm with one specific group of hyperparameters. We draw the Pareto front of $\mathcal{B}^4$ by LOWESS \cite{lowess}.}
     \figvsbottom
    \label{fig:main}
\end{figure*}

\begin{table}[!tb]
    \centering
\resizebox{\linewidth}{!}{
\setlength{\tabcolsep}{2pt}
\begin{tabular}{l| cc |cc |cc |cc }
\toprule
~ & \multicolumn{2}{c|}{Unigram} & \multicolumn{2}{c|}{KGW} & \multicolumn{2}{c|}{AAR} & \multicolumn{2}{c}{EXP}\\
Method & @70 & @80 & @70 & @80 & @65 & @70 & @75 & @80 \\
\hline
~ & \multicolumn{8}{c}{Llama-3-70B} \\
\hline
$\mathcal{B}^4$ & \textbf{30.19} & 80.91 & \textbf{49.11} & \textbf{66.12} & \textbf{75.69} & \textbf{80.60} & \textbf{69.42} & \textbf{75.23} \\
DIPPER & 81.92 & 88.46 & 68.00 & 81.14 & 82.53 & 87.37 & 77.68 & 80.68 \\
RP & 59.77 & \textbf{79.45} & 67.14 & 76.88 & 80.34 & - & 70.69 & 82.92 \\
Base & 72.98 & 83.57 & 67.56 & 81.54 & 83.98 & 89.04 & 73.88 & 78.70 \\
\hline
~ & \multicolumn{8}{c}{Llama-2-7B} \\
\hline
$\mathcal{B}^4$ & \textbf{68.67} & \textbf{89.41} & \textbf{62.38} & \textbf{69.24} & \textbf{73.81} & \textbf{82.69} & \textbf{53.84} & \textbf{69.37} \\
DIPPER & 87.36 & 93.13 & 77.97 & 81.83 & 86.84 & 89.16 & 66.79 & 79.33 \\
RP & 86.91 & 93.51 & 71.67 & 74.06 & 80.89 & 82.99 & 73.57 & 77.85 \\
Base & 82.80 & 91.48 & 73.26 & 78.45 & 86.25 & 88.22 & 67.00 & 75.71 \\
\hline
~ & \multicolumn{8}{c}{Qwen2-72B} \\
\hline
$\mathcal{B}^4$  & \textbf{26.77} & \textbf{54.04} & \textbf{47.53} & \textbf{70.19} & \textbf{60.24} & \textbf{71.15} & \textbf{58.49} & \textbf{68.66} \\
DIPPER & 80.15 & 91.89 & 72.28 & 84.95 & 84.34 & 90.96 & 76.62 & 78.39 \\
RP & 65.16 & 79.47 & 72.64 & 78.67 & 82.36 & 85.93 & 74.34 & 74.34 \\
Base & 84.00 & 84.00 & 75.72 & 75.72 & 83.03 & 87.05 & 76.27 & 76.27 \\
\bottomrule
\end{tabular}
}
\caption{AUC@f ($\downarrow$) of watermark scrubbing attack methods against different victim LLMs protected by different watermark algorithms. The best performance under each setting are \textbf{Bolded}. Dash ('-') indicates an unattainable fidelity threshold for a specific method. 
Note that the selected threshold $f$ varies due to the varying performance of attack methods against different watermarking algorithms.
}
\figvsbottom
\label{tab:main}
\end{table}



\vspace{-1mm}
\subsubsection{Model Choice for Proxy Distributions}
\label{sec:exper-setting-proxy}
\vspace{-1mm}
Our proposed $\mathcal{B}^4$ requires a local available LLM $\theta$ to fit the watermark distribution $\Wdist$ and a paraphraser $\phi$ as $\hFdist$. The choice of $\theta$ is diverse. We consider two possible scenarios in the real life regarding to whether we have access to a model belonging to the same family of the victim LLM. Some leading companies such as Google and Mistral, provides API service of large-scale models, while also open-sourcing some small-scale models with the same architecture. In such scenario, we can surely leverage the small size model as $\theta$ for better distillation performance. Therefore, we apply Qwen2-0.5B to fit $\Wdist$ of Qwen2-72B to explore our method under this setting. In contrast, we apply the Gemma-2-2B-it and Gemma-2-2B to fit $\Wdist$ of Llama-2-7B and Llama-3-70B respectively. The diversity of proxy distributions aims at proving the model-agnostic of our proposed $\mathcal{B}^4$. The specific choices of $\theta$ and $\phi$ are listed in Table \ref{tab:model-setting}.
 More experimental details are presented in Appendix \ref{app-setting}.

\begin{table}[tb]
\small
\centering
\begin{tabular}{ccc}
\toprule
\multicolumn{1}{c}{Victim} & \multicolumn{1}{c}{$\theta$} & \multicolumn{1}{c}{$\phi$} \\ 
\midrule
Llama-2-7B & Gemma-2-2B & Gemma-2-2B-it \\
Llama-3-70B & Gemma-2-2B & Gemma-2-9B-it \\
Qwen2-72B & Qwen2-0.5B & Qwen2-7B-it \\ 
\bottomrule
\end{tabular}
\caption{Selection of the proxy watermark distribution model $\theta$ and the proxy fidelity distribution model $\phi$ against different victim LLMs.}
\figvsbottom
\label{tab:model-setting}
\end{table}

\subsecvs
\subsection{Main Results}
\subsecvs
The primary results of our experiments are illustrated in Figure \ref{fig:main}, with a detailed numerical comparison provided in Table \ref{tab:main}, where we selected several specific fidelity thresholds. Data points with less than 60\% P-SP or more than 95\% ROC-AUC were excluded, as they reflect either poor fidelity or weak efficacy.

\subsecvs
\paragraph{Trade-off between fidelity and efficacy.} Our experimental results across various watermarks and victim models reveal a clear trade-off between fidelity and efficacy, following a distinctive curve that moves from the upper right to the lower left, which aligns with our intuition from Section \ref{sec:exper-metric}. Furthermore, it is worth noting that all attack methods struggle to remove watermarks when the fidelity constraints are stringent.
This also aligns with our intuition. 
The lexical space for modification is rather limited under stringent fidelity constraint. As a result, the output spaces of different attack methods overlap significantly, leading to comparable performance in terms of efficacy.

\subsecvs
\paragraph{\textbf{$\mathcal{B}^4$} outperforms baselines across all settings.} 
Table \ref{tab:main} shows that $\mathcal{B}^4$ consistently achieves the best performance: under the same fidelity constraints, B4 is more effective in scrubbing watermark, reflected by lower ROC-AUCs. 
The superior performance of $\mathcal{B}^4$ is even more evident in Figure \ref{fig:main}, where the Pareto front for $\mathcal{B}^4$ consistently lies below the data points of other methods.
\subsecvs
\paragraph{Robustness of different watermarking methods.} 
An important goal of adversarial attacks is to assess the robustness of various watermarking algorithms. We observe that the other three attack baselines generally struggle to effectively remove watermarks and have difficulty differentiating the robustness of different watermarking techniques. 
However, with its superior efficacy, $\mathcal{B}^4$ is able to expose the differences among various watermarking schemes.
Our results indicate that $\mathcal{B}^4$ is most efficient in scrubbing Unigram, followed by EXP and KGW, while AAR is the most difficult to remove. This may be related to the learnability of the watermarking patterns. For example, Unigram is easier to learn because it maintains a fixed green-red word list, whereas learning AAR watermark requires learning the sampling distribution pattern based on a context window, making it more challenging. 
\subsecvs
\paragraph{Comparing different models.}
In Section \ref{sec:exper-setting-proxy}, we introduced two proxy watermark distribution model settings for $\mathcal{B}^4$: using smaller models from either the same or a different family as the victim model.
Our experiments reveal that using small models from the same family (e.g., attacking Qwen2-72B with Qwen2-0.5B) serves as a more effective proxy. This finding aligns with our intuition, as models from the same family share an identical tokenizer and training data, enabling them to more easily capture the watermarked patterns during distillation, instead of struggling with the domain shifts.


\subsecvs
\subsection{Further Analysis}

\begin{figure}[tb]
    \centering
    \includegraphics[width=\linewidth]{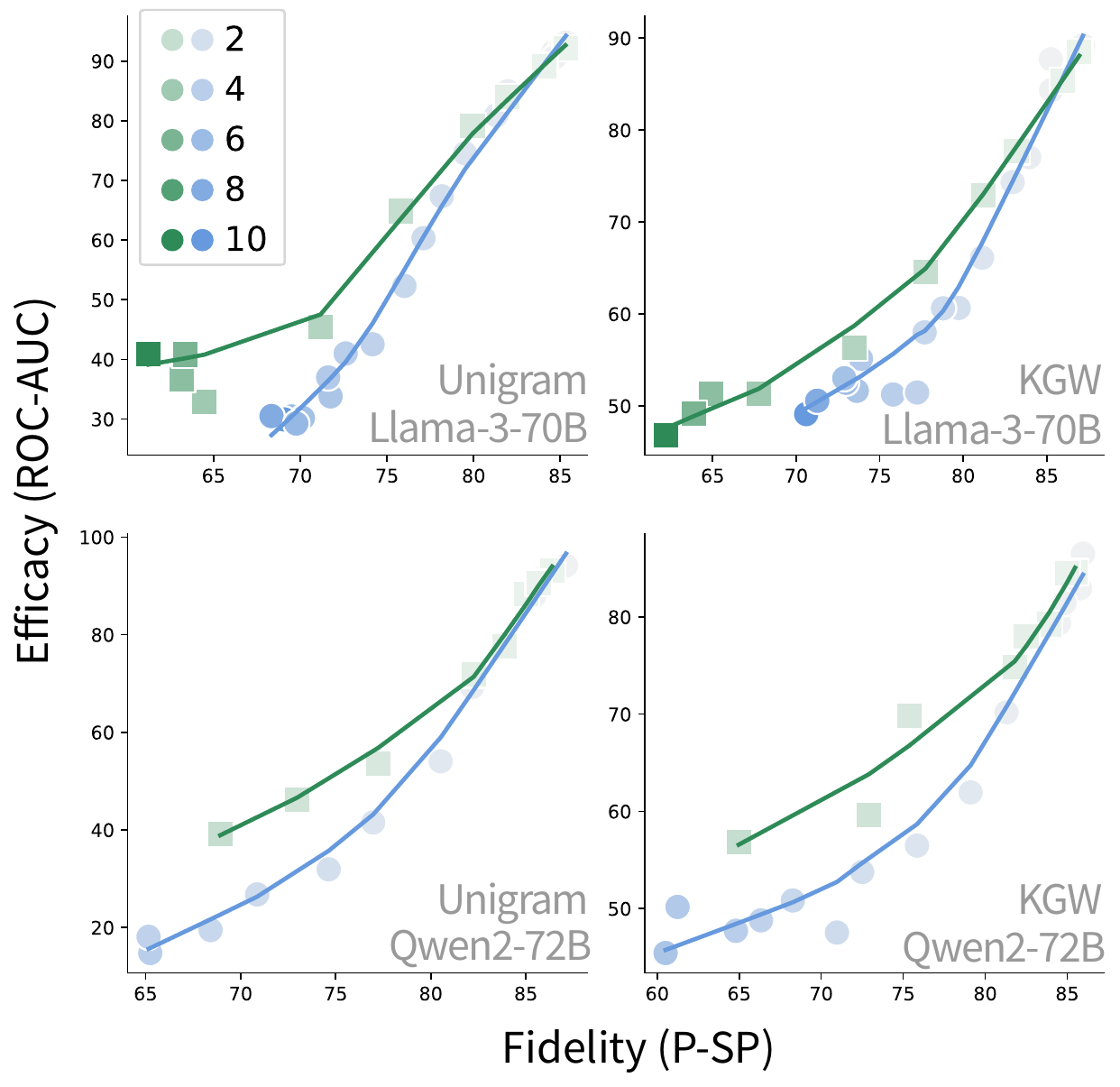}
    \caption{Ablation of $\epsilon$ and AEA. The blue dots and lines denote performance of $\mathcal{B}^4$ with AEA while the green denotes that without AEA. The value of $\epsilon$ is indicated by the shading of dots, with darker colors representing larger $\epsilon$.}
    \figvsbottom
    \label{fig:hyper-ablation}
\end{figure}

\subsubsection{Ablation Study}



Here we explore the effect of hyperparameter $\epsilon$ and the approximation error adjustments (AEA), as shown in Figure \ref{fig:hyper-ablation}.
\subsecvs
\paragraph{Effect of $\epsilon$.} With the growth of $\epsilon$, semantic fidelity decreases while attacking efficacy increases smoothly. This phenomenon aligns with the role of $\epsilon$ as the level of fidelity constraint in Problem \ref{eq:problem}, further justifying our optimization problem.
\subsecvs
\paragraph{Effect of AEA.} 
It is evident that approximation error adjustment consistently yields performance improvements to $\mathcal{B}^4$, which becomes more significant under lower fidelity constraints.
Notably, without of AEA, an unusual phenomenon arises when attacking Unigram-protected Llama-3-70B: as $\epsilon$ increases to a large value, the curve deviates abnormally, shifting towards the upper left, indicating both diminished fidelity and efficacy.
But this abnormal phenomenon disappears after the adjustment is applied, with fidelity and efficacy exhibiting a normal trade-off relationship.
It suggests that the $Q^*$ calculated without AEA is no longer local optimal, which further implies the approximation error discussed in Section \ref{sec:aea} may have greater influence when the constraint in Problem \ref{eq:problem} is loose. 
These findings again underscore the necessity of our proposed adjustments.

\begin{figure}[!tb]
    \centering
    \includegraphics[width=\linewidth]{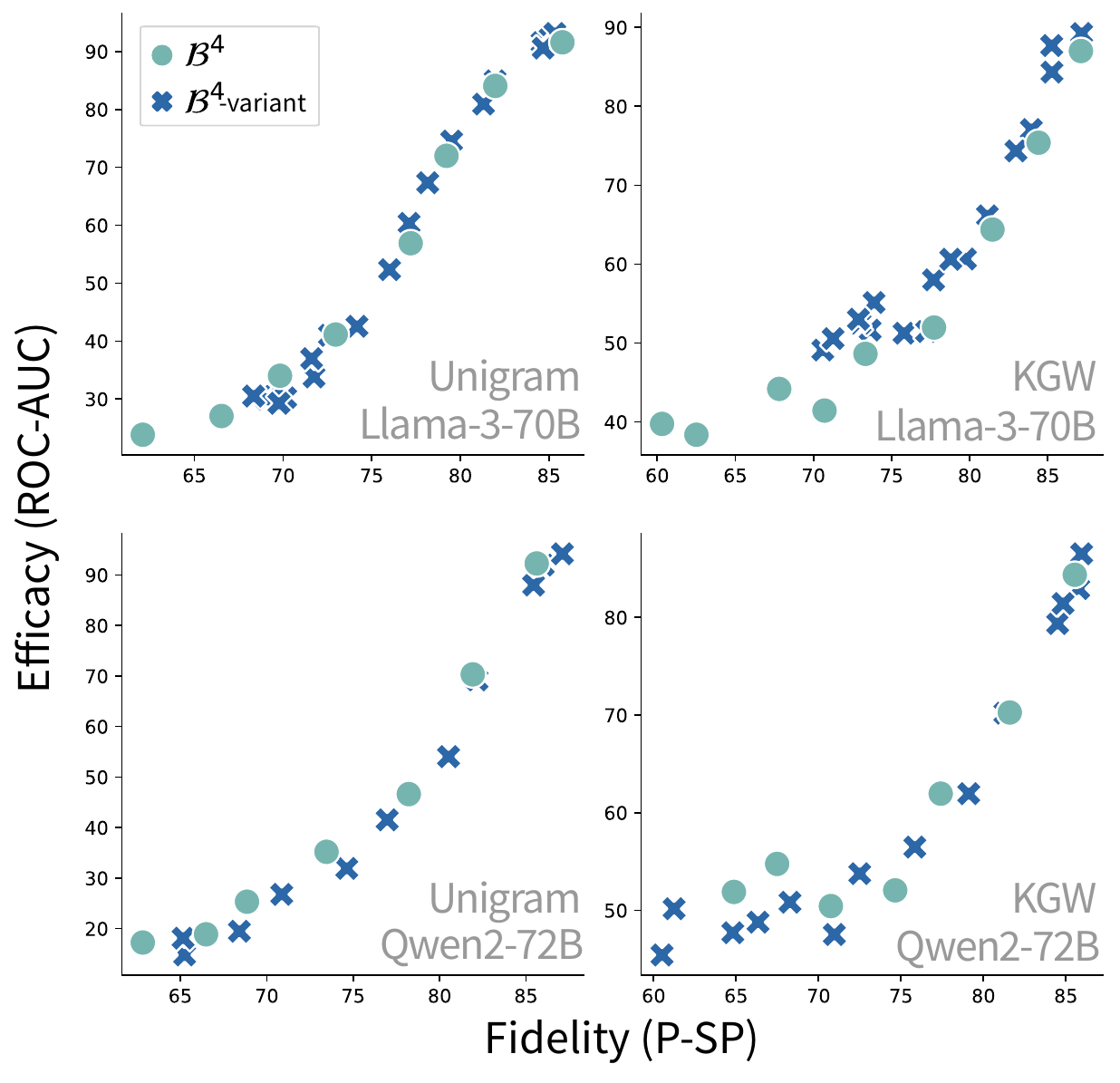}
    \caption{Comparison between $\mathcal{B}^4$ and $\mathcal{B}^4$-variant.}
    \figvsbottom
    \label{fig:closed_form}
    \figvsbottom
\end{figure}

\subsecvs
\subsubsection{$\mathcal{B}^4$-Variant for Speeding Up}
\subsecvs

As discussed in Section \ref{sec:method-problem}, Lagrange multiplier $\lambda^*$ can be determined by solving the equation $\KL(Q^*||\Fdist)=\epsilon$. In other words, $\lambda^*$ is implicitly dependent on the hyper-parameter $\epsilon$ via an implicit function $\lambda^*=\Lambda(\epsilon)$. Therefore, instead of parameterize $\epsilon$, we can parameterize $\lambda^*$, bypassing the need for solving the equation. We refer to this speedup approach as $\mathcal{B}^4$-variant, and we compare its performance to the original parametrization version in Figure \ref{fig:closed_form}. Empirical results show that the speedup variant performs on par with the original $\mathcal{B}^4$, offering a comparable solution with reduced complexity.

\subsecvs
\subsubsection{On the Scale of Training Corpus}
\subsecvs

 We apply 200,000 training samples to distill the proxy watermark distribution $\hWdist$, which is at a cost of about \$90\footnote{The price is calculated based on OpenAI's ChatGPT API pricing (gpt-3.5-turbo-0125).}.
 Here we further discuss how the size of training dataset effects the performance of our method. We start by exploring this question: how many training samples are sufficient for a proxy model to learn the watermark distribution? 
 We utilize Gemma-2-2B and Qwen2-0.5B as proxy models to fit the watermark distribution of Llama-3-70B and Qwen2-72B, respectively, on different scales of watermarked training corpus.
 As shown in Table \ref{tab:ablation-data}, when the training corpus shrinks from 250,000 to 100,000, we don't see a noticeable ROC-AUC drop of proxy models --- With only 100,000 watermarked samples, a small proxy model is already able to learn the watermark distribution pretty well.

\begin{figure}[tb]
    \centering
    \includegraphics[width=\linewidth]{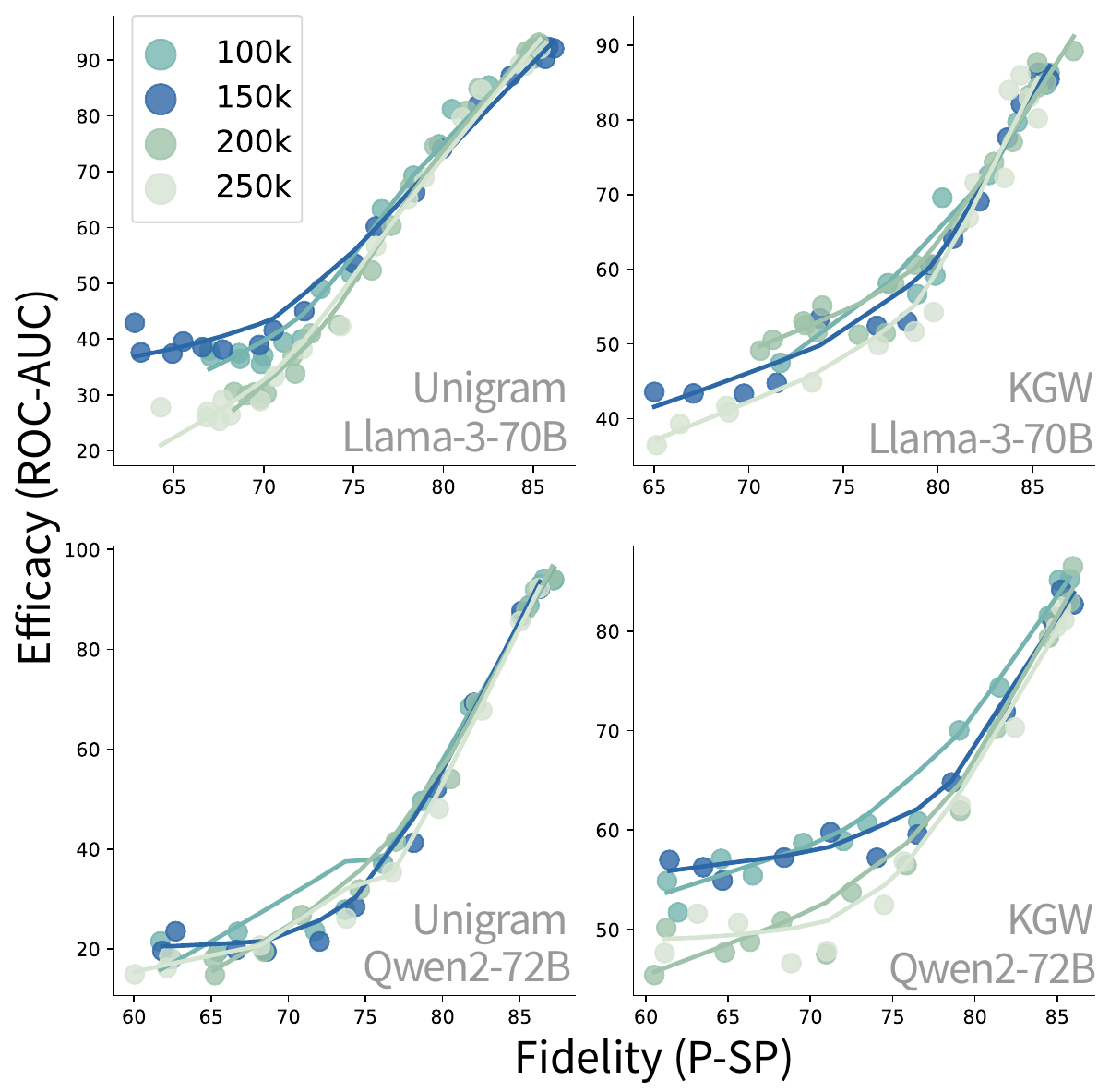}
    \caption{Performance of $\mathcal{B}^4$ with different size of training corpus for approximating watermark distribution.}
    \label{fig:training_data_size}
    \figvsbottom
\end{figure}

We present the performance of our $\mathcal{B}^4$ method under different training dataset settings in Figure \ref{fig:training_data_size}. We find that a dataset of 200,000 watermarked samples are sufficient for training, and 100,000 is also enough for efficient scrubbing attacks. 

\begin{table}[!t]
\centering
\resizebox{\linewidth}{!}{
\begin{tabular}{@{}c|cc|cc}
\toprule
 & \multicolumn{2}{c|}{Qwen2-72B} & \multicolumn{2}{c}{Llama-3-70B} \\
\multicolumn{1}{c|}{Training Data} & \multicolumn{1}{c}{KGW} & \multicolumn{1}{c|}{Unigram} & \multicolumn{1}{c}{KGW} & \multicolumn{1}{c}{Unigram} \\ \midrule
100,000 & 99.17 & 100.00 & 96.79 & 98.83 \\
150,000 & 98.97 & 100.00 & 99.02 & 98.44 \\
200,000 & 99.15 & 100.00 & 98.57 & 99.01 \\
250,000 & 99.24 & 99.98 & 98.57 & 99.46 \\ \bottomrule
\end{tabular}}
\caption{ROC-AUC ($\uparrow$) of proxy models fitting to Unigram/KGW watermarked Qwen2-72B and Llama-3-70B with different sizes of watermarked training dataset.}
\label{tab:ablation-data}
\figvsbottom
\end{table}

\secvs
\section{Related Work}
\subsecvs

\paragraph{LLM Watermarking.}
Early foundational work in LLM watermarking include KGW \cite{kirchenbauerWatermarkLargeLanguage2023} and AAR \cite{aaronson-watermark}, both of which have significantly influenced subsequent research. Following KGW, various methods have been proposed to enhance performance by introducing innovations such as alternative hash functions \cite{kirchenbauerReliabilityWatermarksLarge2023,houSemStampSemanticWatermark2023,liuSemanticInvariantRobust2023,renRobustSemanticsbasedWatermark2023}, heuristic partitioning strategies \cite{liImprovingGenerationQuality2023,chenXMarkLosslessWatermarking2023}, multi-bit message embedding \cite{wangCodableTextWatermarking2023}, and more robust hypothesis testing techniques \cite{fernandezThreeBricksConsolidate2023}.
\subsecvs
\paragraph{Watermark Scrubbing Attack.}
Several previous studies have explored adversarial attacks against watermarking algorithms \cite{sadasivanCanAIGeneratedText2023, pangAttackingLLMWatermarks2024}. Both \citet{jovanovicWatermarkStealingLarge2024} and \citet{zhangLargeLanguageModel2024} focus on stealing the green-list vocabulary in KGW by analyzing token frequency statistics, which requires prior knowledge of the context window size for accurate estimation. \citet{zhangLargeLanguageModel2024} further formalize the task of stealing the green-list vocabulary into a mixed-integer programming problem, though this method relies on access to the full vocabulary and tokenizer, making it difficult to apply in real-world black-box settings.
Another line of work aims to develop black-box scrubbing attacks. Both \citet{zhangWatermarksSandImpossibility2023} and \citet{sadasivanCanAIGeneratedText2023} propose recursive paraphrasing pipelines, while \citet{krishnaParaphrasingEvadesDetectors2023} introduce DIPPER, a paraphraser specifically tuned to evade AI-generated text detection.
Among all, our work is most closely related to the findings of \citet{gu2024learnability}, who demonstrate that model distillation is an effective technique for watermark spoofing. This insight has motivated the design of our proxy watermark distribution. 
\secvs
\section{Conclusion}
\secvs
In this work we research into the watermark scrubbing attack method within the black-box setting, a practically significant but under-studied field. Without needing to know details or even the type of watermarking method used, we derive the format of local-optimal adversarial sample distribution by approximating a fidelity distribution and a watermark distribution. Our proposed attack approach, $\mathcal{B}^4$, can effectively remove the watermark pattern without distorting the original semantic fidelity, demonstrating its superior performance over all baseline models across a wide range of victim settings.

\section*{Limitations}
In this work, we apply the basic distillation technique to approximate the watermark distribution. 
However, the watermark distillation is quite different from normal distillation, since we aims at learning a specific token pattern, instead of learning knowledge.
We do not discuss a distillation technique specific designed for watermark distillation due to space limitations. We will explore this further in future work.

\section*{Ethical Statements}

This work explores adversarial attacks against proprietary LLM watermarking protections. While we recognize the potential implications, we are unaware of any current real-world deployments of watermarking techniques that could be impacted by our methods. Therefore, we believe the risks of malicious applications are limited at this time, and our research cannot be exploited in practice under present conditions. Nonetheless, we emphasize that our primary goal is to advance the understanding of watermark robustness, contributing to the development of more secure and resilient systems.

\bibliography{custom,attack,wm}

\appendix
\clearpage
\section{Preliminary on Watermarking Methods}
\label{app:watermarks}
In this section we introduce the four watermaring methods used in our work.
\begin{itemize}
        \item KGW \cite{kirchenbauerWatermarkLargeLanguage2023}. In each decoding step, KGW utilizes a $c$-length context window $\rvy_{i-c:i-1}$ as a random seed to partition the vocabulary $\Sigma$ into a green list and a red list. The size of the green list is controlled by a hyperparameter $\gamma$. A perturbation $\tau$ is then applied to the logits of these green-list tokens to shape the watermark distribution. 
        \item AAR \cite{aaronson-watermark}. AAR also applies a $c$-length context window as a seed to generate a standard Gumbel distribution vector, which is then utilized for Gumbel-max sampling \citep{debreu1960individual}. Essentially, the final watermark distribution becomes a degenerate distribution based on the sampled token.
        \item Unigram \cite{zhaoProvableRobustWatermarking2023}. Similar to KGW, Unigram partitions the vocabulary but eliminates the dependency on the context window. Instead, it maintains a fixed green list throughout the generation process, providing increased robustness against adversarial attacks.
        \item EXP \cite{kuditipudiRobustDistortionfreeWatermarks2023}. EXP also borrows the idea of Gumbel-max sampling in AAR. However, instead of seeding on the context window, it maintains a global private seed sequence, where each seed corresponding to a specific time step during decoding.
\end{itemize}

\section{Pseudocodes for $\mathcal{B}^4$}
\label{app-pcodes}

\begin{algorithm}[H]
\caption{Watermark Distribution Approximation}
\KwIn{Training corpus $D$, watermark distribution $\Wdist$, initialization weight $\theta_{ini}$}
\KwOut{Proxy watermark distribution $\hWdist$}

$\theta \gets\theta_{ini}$\;

\ForEach{epoch}{
    \ForEach{$\rvy \in D$}{
        $\mathcal{L}(\theta) \gets -\sum_i \log p_\theta(\rvy_i | \rvy_{<i})$\;
        
        $\theta\gets\theta-t\nabla{L}(\theta)$
    }
}

$\hWdist \gets p_\theta$\;

\Return{$\hWdist$}\;
\end{algorithm}

\begin{algorithm}[H]
\small
\caption{$\mathcal{B}^4$ with Approximation Error Adjustment}
\KwIn{Watermarked sample $\rvy^w$, proxy watermark distribution $\hWdist$, proxy model after distillation $\theta$, proxy model before distillation $\theta_{ini}$, paraphraser distribution $\hFdist$, paraphraser $\phi$, hyperparameter $\mu$}
\KwOut{Scrubbing attack distribution $Q^*$}

\ForEach{decoding step $i$}{
    
    $\Sigma_u^i \gets \{v \in \Sigma_i : |p_\theta(v | \rvy_{<i}) - p_{\theta_{ini}}(v | \rvy_{<i})| < \mu\}$\;
    
    $\lambda^* \gets \texttt{solve\_lagrange\_multiplier}(\hFdist_{;\phi},\hWdist_{;\theta},\Sigma-\Sigma_u^i)$\;

    \ForEach{$\rvy_i \in \Sigma_u^i$}{
        $Q^*(\rvy_i | \rvy_{<i}, \rvy^w) \gets \hFdist(\rvy_i | \rvy_{<i}, \rvy^w; \phi)$\;
    }
    \ForEach{$\rvy_i \in \Sigma-\Sigma_u^i$}{
        $Q^*(\rvy_i | \rvy_{<i}, \rvy^w) \gets \frac{\hFdist^{\frac{1}{1-\lambda}}(\rvy_i | \rvy_{<i}, \rvy^w; \phi)}{\hWdist^{\frac{\lambda}{1-\lambda}}(\rvy_i | \rvy_{<i}; \theta)}$\;
    }
}
\Return{$Q^*$}\;

\SetKwFunction{FMain}{solve\_lagrange\_multiplier}
\SetKwProg{Fn}{Function}{:}{}
\Fn{\FMain{$\hFdist_{;\phi},\hWdist_{;\theta},\Sigma$}}{
    $\lambda \gets \lambda_0$\; 
    
    $h(\lambda,v) \gets \texttt{lambda }\lambda,v: \frac{\hFdist^{\frac{1}{1-\lambda}}(v | \rvy_{<i}, \rvy^w; \phi)}{\hWdist^{\frac{\lambda}{1-\lambda}}(v | \rvy_{<i}; \theta)}\;$
    
    $f(\lambda) \gets \texttt{lambda }\lambda:  \sum_{v\in\Sigma}h(\lambda,v)\log\frac{h(\lambda,v)}{\hFdist(v)}-\epsilon\;$
    
    \While{$|f(\lambda)| < 10^{-6}$}{
        $\lambda \gets \lambda - \frac{f(\lambda)}{f'(\lambda)}$\;
    }
    $\lambda^*\gets\lambda\;$
    
    \Return{$\lambda^*$}
}
\end{algorithm}

\section{Experimental Details}
\label{app-setting}

\paragraph{Watermark algorithms.}
We implement the four watermarking algorithms mentioned above. For both KGW and Unigram, we apply a common setting with $\gamma=0.5, \tau=2$ following \citet{kirchenbauerWatermarkLargeLanguage2023}. For EXP, we maintain a key sequence of length $256$, enough to seed the evaluation datasets of token length $200$. For both KGW and AAR, we set the context width $c$ to 1 for better robustness.

\paragraph{Corpus Construction.}
We randomly select 200,000 samples from the English(en) subset of the C4 dataset and truncate each sample to the first 50 tokens to create the prompting dataset. We then apply each of the four watermark methods, i.e., KGW, Unigram, AAR and EXP, to query the victim models with these 50-token prompts, to generate 200,000 responses of 200-tokens. This results in a training corpus containing responses from a victim LLM embedded with a specific type of watermark.

\paragraph{Training Details.}
We then fine-tune the proxy models on each training corpus for 5 epochs, with a batch size of 128 sequences, sequence length of 256 tokens. We save the checkpoint of each training epoch and only reserve the one with the best validation result. We follow \citet{gu2024learnability} to set the maximal learning rate to 1e-5, and use cosine learning rate decay with a linear warmup for the first 500 steps. Training a Llama-2-2B proxy model took approximately 9 hours on 4 NVIDIA RTX A6000 48 GPUs, and training a Qwen2-0.5 proxy model took approximately 4 hours on 1 NVIDIA RTX A6000 48 GPU.

\paragraph{Hyperparameter Space.} To generate a comprehensive Pareto front, we enumerate several sets of hyperparameters for each baseline method. The hyperparameter spaces for the different methods are listed in Table \ref{tab:space}. Since the data points corresponding to these baselines are not uniformly distributed in the Fidelity-Efficacy plot across different experimental settings, we manually add some specific hyperparameter settings where necessary to ensure accurate representation.

\begin{table}[bt]
    \centering
    \small
    \begin{tabular}{c|c}
    \toprule
    \multicolumn{2}{c}{RP} \\
    \midrule
       paraphrase counts  & [1,2,3,4,5] \\
       chunk size  & [1,2]\\
       prompt & Instruction 1 in Table \ref{tab:prompts}\\
    \midrule
    \multicolumn{2}{c}{DIPPER} \\
    \midrule
       lex  & [30,35,40,45,50,55,60,65,70] \\
    \midrule
    \multicolumn{2}{c}{Base} \\
    \midrule
       beam size  & [1, 10] \\
       prompts  & 12 instructions in Table \ref{tab:prompts} \\
    \midrule
    \multicolumn{2}{c}{$\mathcal{B}^4$} \\
    \midrule
        $\epsilon$ & [0.01,0.1,0.5,1,1.5,2,3,4,5,6,10] \\
        beam size & 10\\
        prompt & Instruction 1 in Table \ref{tab:prompts}\\
    \bottomrule
    \end{tabular}
    \caption{Hyperparamter space of different baselines}
    \label{tab:space}
\end{table}

\section{Prompt Instructions}
Here we list instructions we used for paraphrasing:

\makeatletter
\newcommand{\DrawLine}{%
  \begin{tikzpicture}
  \path[use as bounding box] (0,0) -- (\linewidth,0);
  \draw[color=black!75!black,dashed,dash phase=2pt]
        (0-\kvtcb@leftlower-\kvtcb@boxsep,0)--
        (\linewidth+\kvtcb@rightlower+\kvtcb@boxsep,0);
  \end{tikzpicture}%
  }
\makeatother

\begin{tcolorbox}[title = {Instructions for Paraphrase},fontupper=\scriptsize, fontlower=\scriptsize]
\label{tab:prompts}
\textbf{Intruction 1}: Paraphrase the following paragraphs line by line. Don't output any other information except the paraphrased texts. This is the text: {}\\
\DrawLine \\
\textbf{Intruction 2}: You are an expert copy-editor. Please rewrite the following text in your own voice and paraphrase all sentences. Ensure that the final output contains the same information as the original text and has roughly the same length.   Do not leave out any important details when rewriting in your own voice. This is the text:  {}\\
\DrawLine \\
\textbf{Intruction 3}: As an expert copy-editor, please rewrite the following text in your own voice while ensuring that the final output contains the same information as the original text and has roughly the same length. Please paraphrase all sentences and do not omit any crucial details. Additionally, please take care to provide any relevant information about public figures, organizations, or other entities mentioned in the text to avoid any potential misunderstandings or biases. {}\\
\DrawLine \\
\textbf{Instruction 4}: As an expert copy-editor, please rewrite the following text in your own voice while ensuring that the final output contains the same information as the original text and has roughly the same length. Please paraphrase all sentences and do not omit any crucial details. Don't output any other information except the paraphrased texts. This is the text: {} \\
\DrawLine \\
\textbf{Intruction 5}: Paraphrase the following paragraphs line by line. Try to keep the similar length to the original paragraphs. Don't output any other information except the paraphrased texts.This is the text: {}\\
\DrawLine \\
\textbf{Intruction 6}: As an expert copy-editor, please rewrite the following text in your own voice while ensuring that the final output contains the same information as the original text and has roughly the same length. Please paraphrase all sentences and do not omit any crucial details. Don't output any other information except the paraphrased texts. This is the text: {} \\
\DrawLine \\
\textbf{Intruction 7}: Paraphrase the following paragraph such that it preserves the original meaning but uses different phrasing and vocabulary. Ensure that the new version has minimal overlap with the original in terms of common phrases, word sequences, and n-grams. Output should be natural, coherent, and maintain the key information from the source text. Here are the texts: {}\\
\DrawLine \\
\textbf{Intruction 8}: Paraphrase the following paragraph such that it preserves the original meaning but uses different phrasing and vocabulary. Ensure that the new version has minimal overlap with the original in terms of common phrases, word sequences, and n-grams. Output should be natural, coherent, and maintain the key information from the source text. Here are the texts: {}\\
\DrawLine \\
\textbf{Intruction 9}: Paraphrase the following paragraph such that it preserves the original meaning but has minimal overlap with the original in terms of common phrases, word sequences, and n-grams. Here is the text: {}\\
\DrawLine \\
\textbf{Intruction 10}: Paraphrase the following paragraph in your own tone. Ensure that it has minimal overlap with the original in terms of common phrases, word sequences, and n-grams. Here is the texts: {}\\
\DrawLine \\
\textbf{Intruction 11}: Rewrite the following paragraph in a way that retains its core meaning but alters its wording and structure. Focus on minimizing shared n-grams and phrases between the original and the rewritten text, while keeping the content clear and coherent. Here are the texts: {}\\
\DrawLine \\
\textbf{Intruction 12}: Transform the following paragraph into a new version that conveys the same message but is expressed with different wording and phrasing. Try to keep n-gram overlaps minimal, employing synonyms, rephrased expressions, and varied sentence patterns. Here are the texts: {}\\
\DrawLine \\
\textbf{Intruction 13}: Create a paraphrased version of the provided text such that it maintains the semantic essence while minimizing the similarity in wording and n-gram patterns. Focus on using distinct phrases and vocabulary to achieve a high degree of linguistic diversity. Here are the texts: {}\\
\end{tcolorbox}

\clearpage
\section{Sketch for Optimization Problem Solution}

In this section, we present the solution process of the proposed optimization problem \ref{eq:problem}.

We begin with formally introducing the Slater Constraint Qualification for Convex Inequalities \cite[Proposition 3.3.9]{bertsekas1999nonlinear}
\begin{lemma}
    Let $x^*$ be a local minimum of the problem
    \begin{align*}
        \min\; &f(x)\\
        s.t.\quad &g_i(x)\leq0,\,\forall i\\
        &h_j(x)=0,\,\forall j\\
    \end{align*}
    ,where $f$ and $g_i$ are continuously differentiable.
    Assume that $h_j$ are linear, $g_i$ are convex and there exists a feasible vector $x_0$ satisfying
    \begin{equation*}
        g_i(x_0)<0,\,\forall i
    \end{equation*}
    Then $x^*$ satisfies the KKT conditions.
\end{lemma}

In our proposed optimization problem, the objective function $f=-\KL(Q||\Wdist)$ and the inequality $g=\KL(Q||\Fdist)-\epsilon$ in Problem \ref{eq:problem} involves KL-divergence with the range in $[0, \infty)$. And the equality $h=\sum_\rvy Q(\rvy)-1$ is linear. We immediately have the Slater Constraint Qualification satisfied.

Now that we can derive the format of local minimum point $Q^*$ by solving KKT conditions. Based on the Stationarity Condition, for all $\rvy\in\Sigma^*$, we have 
{
\thinmuskip=1mu
\medmuskip=0.5mu
\thickmuskip=0.5mu
\begin{gather*}
    \frac{\partial L}{\partial Q^*(\rvy)}=0\\
    -(\log\frac{Q^*(\rvy)}{\Wdist(\rvy)}+1) +\lambda^* (\log\frac{Q^*(\rvy)}{\Fdist(\rvy|\rvy^w)}+1)=0\\
    Q^*(\rvy)\propto\Fdist^{\frac{1}{1-\lambda^*}}(\rvy|\rvy^w)\Wdist^{-\frac{\lambda^*}{1-\lambda^*}}(\rvy)
\end{gather*}}

Further with Complementary Slackness, we have either $\lambda^*=0$ or $\KL(Q^*||\Fdist)=\epsilon$. The former equation indicates a trivial solution, where $Q^*=\Fdist$. Therefore, we focus on the latter equation, 
which can be numerically solved via the commonly used Newton-Raphson Method.

\end{document}